# Tifinagh-IRCAM Handwritten character recognition using Deep learning

El Wardani DADI

SDIC team, Laboratory of Applied Sciences,
National School of Applied Sciences of Al-Hoceima (ENSAH)
University of Abdelmalek Essaadi
w.dadi@uae.ac.ma

**Absract.**

In this paper, we exploit the benefits of the deep learning approach to design an efficient system of Amazigh handwritten recognition. Indeed, this approach has proved a greater efficiency in the various domains, especially recognition tasks. However, to take full advantage of this approach it's necessary to construct an adequate dataset of training and testing that represent faithfully the concerned problem. To this end, we have prepared our dataset of 102 writers each one contains 33 characters of IRCAM-Tifinagh. Inspired by the MNIST database, the set of characters is size-normalized and centered in a fixed-size image. The r*esulting is a* grey level image of size 28x28, where the black color is the non-color of the character. The number of images produced after this preprocessing step is 3,366.

**Keysword**: handwritten, deep learning, CNN, OCR, recognition.

## 1. Introduction

Handwriting recognition is an active research area in pattern recognition and computer vision. This domain has several applications fields such as documents digitizing and archiving, postal address interpretation, signature verification.... Currently, many Smartphone applications are based on handwritten as input by touching the screen through a stylus or finger. These applications are used for example to take notes.

In recent years, this domain has received much attention in academic areas and it has become one of the most popular research subjects because of developing the new classification and recognition approach. Indeed, the first artificial pattern recognizers system to achieve human competitive performance was performed by Yann LeCun et al[5], on the famous MNIST handwritten digits dataset. This system is well known in the literature by Convolutional Neural Network(CNN) which is a kind of deep learning approach. This last has proven very interests and effective results in classification and retrieval tasks.

In this paper, we are interested in Tifinagh character which is the Amazigh language alphabet. Indeed, the Amazigh language is widely used in North Africa in countries such as Morocco, Algeria, Libya, Tunisia, Mauritania, etc. This language and its alphabet Tifinagh have been recently recognized. The Tifinagh code approval by the International Organization for Standardization (ISO) that was in 2004 has allowed its use in several computer systems nationally and internationally. The Tifinagh writing is currently integrated by major companies specializing in the production of software, which will facilitate its use especially in terms of Internet and text processing.

The emergence of Tifinagh character in the digital world has created an increased need for several specific computer systems like those developed for other languages, namely translation systems, automatic document filing systems, security systems, automatic processing of administrative documents, postal automation,... These systems can use a handwritten character recognition system.

To promote machine learning and pattern recognition research, several standard databases have emerged in which the handwritten digits are preprocessed, including segmentation and normalization, so that researchers can compare recognition results of their techniques on a common basis. This dataset needs to represent a sufficiently challenging problem to make it both useful and to ensure its longevity. For Tifinagh handwritten dataset, at our knowledge two databases are proposed [1, 2]. The two datasets are developed basing on 33 characters of IRCAM-Tifinagh adopted by the Royal Institute of Amazigh Culture, Morocco. These two databases contain an insufficient number of examples and also the examples that do not represent every possible case. Furthermore, the existence of a varied suite of benchmark tasks is important in allowing a more holistic approach to assessing and characterizing the performance of an algorithm or system.

In this context, we propose a new dataset of Tifinagh handwritten. To construct this dataset we have selected a normal sample of 102 writers that represent different age and education levels. To outperform a simple-to-use dataset, our preprocessing operation is inspired by the famous MNIST dataset[3], where the set of characters is size-normalized and centered in a fixed-size image. The resulting is a grey level image of size 28x28, where the black color is the non-color of the character. This is a relatively simple-to-use dataset for people who want to try machine learning techniques and pattern recognition methods, especially the deep learning approach, on real-world data while spending minimal efforts on preprocessing and formatting.

The proposed dataset is used to develop an efficient recognition system basing on deep learning. Indeed, this approach and in particular the convolutional neural networks (CNNs), have proven to be very efficient in the most recognition task.

## 2. Dataset

In this paper, we propose a new pre-processed dataset of Tifinagh handwritten character. Our proposed database is inspired by the famous MNIST dataset, where the set of characters is size-normalized and centered in a fixed-size image. The resulting is a grey level image of size 28x28, where the black color is the non-color of the character. This is a relatively simple-to-use dataset for people who want to try machine learning techniques and pattern recognition methods, especially the deep learning approach, on real-world data while spending minimal efforts on preprocessing and formatting.

To preprocessing this data, we have defined a particular and general treatment. The particular one concern especially the first alphabet which is a = " □ ". Indeed, in Tifinagh writing, this character takes a smaller size than the other alphabets especially on the vertical level (cf. Figure 1).

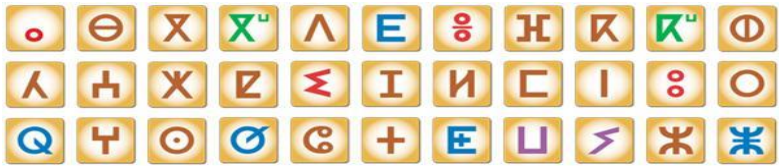

Figure 1: The 33 IRCAM-Tifinagh alphabets

The different pretreatments applied are framing, adjustment and standardization. The main objective is to produce images of the same size (28 * 28), taking inspiration from the MNIST database.

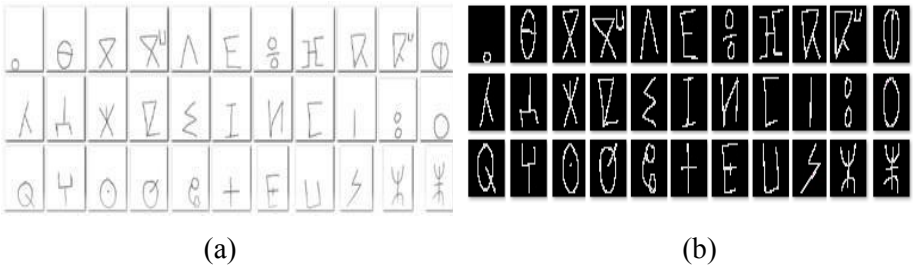

(a)                                                          (b)

Figure 2: IRCAM-Tifinagh handwritten sample: (a) before and (b) after preprocessing

Our preprocessing steps are performed as follow:

- As the characters do not fill the entire image, the region around the actual character is extracted.
- Then it is placed and centered into a square image with the aspect ratio preserved.
- The region of interest is padded with a 2 pixels border when placed into the square image.
- Finally, the image is down-sampled to 28 × 28 pixels using bi-cubic interpolation.

## 3. Training and test using CNN

For the CNN architecture, we have used a simple one constructed by two convolutions layers and 2 layers of max-pooling.

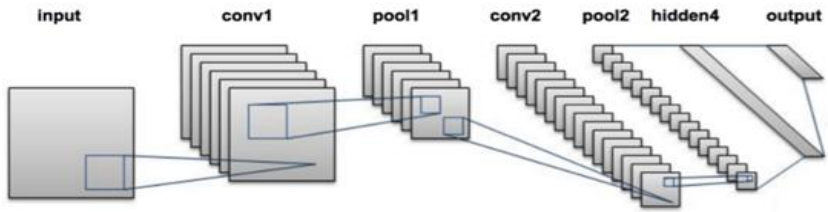

Figure 3: the used CNN architecture.

To perform tests, we have split our dataset as follows: 86% (88 samples) for the train split and 14% for the test split.

The training process is performed on a PC equipped with NVDIA GeForce GT 740. Table 1 and Figure 4 show the accuracy obtained after 100 epochs of training and testing, which is a very important and interesting precision. Indeed, the obtained results for both operations are very close to 100%. In addition, as shown in Figure 4, with more eras, the loss and accuracy of the model on the training and test data converged, thus making the model stable.

Table 1 : Accuracy

|       | Top-1  | Top-5  |
|-------|--------|--------|
| Train | 0.9969 | 1.0000 |
| Test  | 0.9475 | 0.9960 |

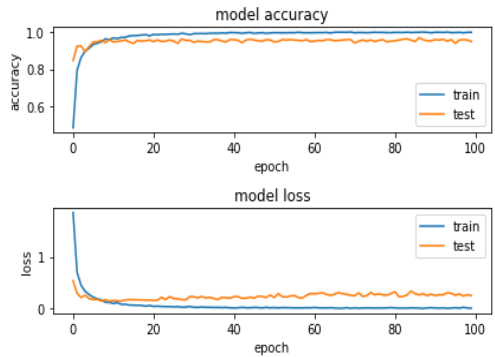

Figure 4: evolution of accuracy and loss for 100 epochs

## 4. Conclusion and perspectives

In this paper we have presented a dataset of IRCAM-Tifinagh handwritten character constructed of 102 samples. This dataset is used to develop a recognition system of the concerned alphabet using deep learning approach. Our recognition system presents an important accuracy.

The number of samples presented in this work still insufficient, so it's crucial to construct a large database. To this end, our solution consists to use the data augmentation. This technique serves not only to have a large database but also to have a system which characterized by the property called invariance. More specifically, a CNN can be invariant to translation, viewpoint, size or illumination (Or a combination of the above). Moreover and for more challenging dataset, in the next work we want to add some other writers as perturbed sample. This perturbation consists of adding characters with missing parts.